\documentclass[12pt]{article}
\usepackage{comment}
\usepackage{amsmath}
\usepackage{amssymb}
\usepackage{graphicx}
\usepackage{here}
\usepackage{mathtools}
\usepackage{subcaption}
\usepackage{url}
\usepackage{mathtools}
\usepackage[sort&compress, numbers, merge]{natbib}
\usepackage{physics}
\usepackage{todonotes}

\usepackage[colorlinks=true, linkcolor=blue, citecolor=blue, urlcolor=black]{hyperref}

\begin{document}

\def\ps{\mathbf{p}}
\def\PS{\mathbf{P}}

\baselineskip 0.6cm

\def\simgt{\mathrel{\lower2.5pt\vbox{\lineskip=0pt\baselineskip=0pt
           \hbox{$>$}\hbox{$\sim$}}}}
\def\simlt{\mathrel{\lower2.5pt\vbox{\lineskip=0pt\baselineskip=0pt
           \hbox{$<$}\hbox{$\sim$}}}}
\def\simprop{\mathrel{\lower3.0pt\vbox{\lineskip=1.0pt\baselineskip=0pt
             \hbox{$\propto$}\hbox{$\sim$}}}}
\def\tr{\mathop{\rm tr}}
\def\SU{\mathop{\rm SU}}

\begin{titlepage}

\begin{flushright}
OU-HET-1138
\end{flushright}
~

\vskip 1.5cm

\begin{center}

{\Large \bf
A unified theory of learning}

\vskip 2cm
{\large
Taisuke Katayose\footnote{\texttt{taisuke.katayose@het.phys.sci.osaka-u.ac.jp}}
}

\vskip 1.0cm
{\it
Department of Physics, Osaka University, Toyonaka 560-0043, Japan
}

\vskip 3.5cm
\abstract{Recently machine learning using neural networks (NN) has been developed, and many new methods have been suggested. These methods are optimized for the type of input data and work very effectively, but they cannot be used with any kind of input data universally. On the other hand, the human brain is universal for any kind of problem, and we will be able to construct artificial general intelligence if we can mimic the system of how the human brain works. We consider how the human brain learns things uniformly, and find that the essence of learning is the compression of information. We suggest a toy NN model which mimics the system of the human brain, and we show that the NN can compress the input information without ad hoc treatment, only by setting the loss function properly. The loss function is expressed as the sum of the self-information to remember and the loss of the information along with the compression, and its minimum corresponds to the self-information of the original data. To evaluate the self-information to remember, we provided the concept of memory. The memory expresses the compressed information, and the learning proceeds by referring to previous memories. There are many similarities between this NN and the human brain, and this NN is a realization of the free-energy principle which is considered to be a unified theory of the human brain. This work can be applied to any kind of data analysis and cognitive science. }

\end{center}

\end{titlepage}

%%%%%%%%%%%%%%%%%%%%%%%%%%%%%%%%
%%%%%%%%%%% Contents %%%%%%%%%%%
%%%%%%%%%%%%%%%%%%%%%%%%%%%%%%%%
\tableofcontents
\newpage

%%%%%%%%%%%%%%%%%%%%%%%%%%%%%%%%%%%%
%%%%%%%%%%% Introduction %%%%%%%%%%%
%%%%%%%%%%%%%%%%%%%%%%%%%%%%%%%%%%%%
\section{Introduction}
\label{sec: Introduction}

Recently machine learning using neural networks (NN) has been developed and many new methods have been suggested \cite{SCHMIDHUBER201585}. For example, convolutional NN\,\cite{DBLP:journals/corr/OSheaN15} for image processing, attention model\,\cite{DBLP:journals/corr/VaswaniSPUJGKP17} for natural language processing, and generative adversarial networks\,\cite{DBLP:journals/corr/SalimansGZCRC16} for classification problem have achieved prominent results. These models are constructed by considering the properties of the input data and reflecting these properties in the model. On the other hand, thinking about the human brain, it does not seem to need such ad hoc treatment and can adapt to any kind of problem. 
Then, what determines how the human brain works, and why is the human brain so versatile? To answer these questions, we need to reconsider the meaning of learning and find out the theory behind it. In physics, there is the principle of least action, and the motion of objects is determined by minimizing the action. There must be a corresponding concept to the theory of learning, and this is nothing but the loss function. Learning proceeds by minimizing the loss function. From this perspective, we consider that the essence of the problem is in the loss function and the NN model is just a tool to find a minimum of the loss function. In other words, the upgrade of the NN model may improve the convergence of the calculation but cannot solve the unsolvable problem because of the bad setting of the loss function. This situation is commonly happening in the study of the NN, so we need not only to make a new NN model for the individual cases but also to reconsider the essential meaning of the loss function. We claim that the loss function is not to be designed for good results but to be defined theoretically. The goal of this paper is to understand the way how the human brain works and to derive the general loss function which enables the NN to work similarly to the human brain by minimizing it. We emphasize that the actual implementation is not discussed in this paper.  

To achieve our goal, we need to define learning mathematically and derive the loss function in the calculable form. By careful consideration, we found that learning is the compression of information. We make a toy NN model and then define the loss function which enables the compression of information. This NN is a kind of autoencoder, and we do not need teacher data or ad hoc treatment for this NN. We referred to the system of the human brain to construct this NN model, and the outstanding property is recording the output from the middle layer every time a new input enters. This is called memory, and the loss function is defined using these memories. The memory denotes the compressed information, and it can express the abstract information as same as actual human memory. We can find many similarities between this NN and the human brain, and this work also can be used in cognitive science.

This paper is arranged as follows. In Sec.\,\ref{sec: Consideration about human learning}, we consider the essential meaning of learning and mention that learning is the compression of information. In Sec.\,\ref{sec: The model of neural network} we define a toy NN model which mimics the system of the human brain. This is an autoencoder type NN model, and we do not go into the details of the model, because our goal is to derive the universal loss function. 
In Sec.\,\ref{sec: The self-information and the loss function}, we define the self-information to remember and loss of information and show that the sum of these quantities corresponds to the self-information of the input data. We also show that the loss function is defined as the prediction of these quantities through the NN. In Sec.\,\ref{sec: The role of the memory and its probability distribution}, we discuss the concept of memory and introduce the concept of the hidden variables. We show that the memory expresses the compressed and abstract information as similar to the memory of the human brain. In Sec.\,\ref{sec: The uncertainty of the solution}, we mention the uncertainty of the solution to minimize the loss function, and we suggest a new loss function to solve this problem. We demonstrate how this loss function works by using simple examples. In Sec.\,\ref{sec: Discussion}, we discuss the reason why our model does not need ad hoc treatment. We also discuss unclear part of the concept of memory and the similarity between this NN and the human brain. In Sec.\,\ref{sec: Conclusion}, we summarize our study.

\section{Consideration about human learning}
\label{sec: Consideration about human learning}
In this section, we consider how the human brain learns things. To make it easy to understand, we discuss the concepts step by step.    

\subsection{General definition of learning}
\label{sec: General definition of learning}
The word learning has a wide range of meaning, and this word is used when we acquire some skills through training. For thinking about learning, we need to define what is learning mathematically. Let us give two examples of learning and find out the common point. For the first example, we can say that a baby learned the word cat when he understands what is a cat. For the second example, we can say that a dog learned tricks, after repeating the training such as feeding him if he follows what the owner says. Then, what is the thing these learning have in common? Our answer to the question is that learning is the discovery of the correlation between information. In the first example, the baby notices that he tends to hear the word cat pronounced when he is seeing a cat, and he discovers the correlation between hearing the pronounced word cat and seeing a cat. In the second example, the dog notice that he can get food after following what the owner says, and he discovers the correlation between having food and following what the owner says. This statement applies to any kind of learning and we can identify the learning as the discovery of the correlation between the information.

\subsection{Correlation and probability bias}
\label{sec: Correlation and probability bias}
Next, let us explain the correlation between the information as discussed in Sec.\,\ref{sec: General definition of learning}. To reveal the essence of the correlation between information, let us explain it using coding theory. Suppose the two bits binary system, which can take the following states defined as
\begin{gather}
\label{eq: s1234}
    E_1 \equiv (0,0),\; E_2 \equiv (0,1),\; E_3 \equiv (1,0),\; E_4 \equiv (1,1). 
\end{gather}
In this system, the first bit takes a 0/1 value and the second bit also takes a 0/1 value, and we consider each bit expresses different information. Making the correspondence to the case with the baby learns the word cat, the first bit expresses whether the baby hears the word cat and the second bit expresses whether the baby sees a cat. In such a case, there emerges the correlation between the first bit and the second bit, and we can think that the probabilities of getting $E_1$ and $E_4$ become higher. This probability bias is caused by the correlation between the information, and we can say that the learning is the detection of the probability bias. 

\subsection{Probability bias and redundancy}
\label{sec: Probability bias and redundancy}
The next question is how the brain detects such probability bias automatically. To clarify what happens when the probability distribution is biased, let us consider the sample case as in Sec.\,\ref{sec: Correlation and probability bias}. We set the probability distribution of the states as follows:
\begin{gather}
\label{eq: case 2}
P(E_1) = 0.6 ,\; P(E_2) = 0.1 ,\; P(E_3) = 0.1,\; P(E_4) = 0.2.
\end{gather}
where $P(\cdot)$ denotes the probability that the argument happens, and we set the probabilities of $E_1$ and $E_4$ higher than others. If the probability is biased, it is known that redundancy emerge. The redundancy $R$ is defined as
\begin{equation}
\label{eq: redundancy}
    R \equiv H_\mathrm{max} - H \, ,
\end{equation}
where $H$ is the information entropy defined as
\begin{equation}
    H \equiv -\sum_{i}P(E_i)\log{P(E_i)} \, ,
\end{equation}
and $H_\mathrm{max}$ is the maximum value of the information entropy when we change the probability distribution. For the case with Eq.\,(\ref{eq: case 2}), $H$ becomes maximum when $E_i$ happens with the same probability, and $H_\mathrm{max}$ is calculated as 
\begin{equation}
\begin{split}
    H_\mathrm{max} &= -\sum_{i=1}^4 (1/4) \log{(1/4)} \\
    &= 1.386\cdots \, .
\end{split}
\end{equation}
The information entropy $H$ is calculated as
\begin{equation}
\begin{split}
    H &= - \sum_{i=1}^4 P(E_i) \log{P(E_i)} \\
    &= 1.088\cdots \, .
\end{split}
\end{equation}
Then the redundancy for this case is calculated as 
\begin{equation}
\begin{split}
    R &= H_\mathrm{max} - H = 0.298 \cdots \, .
\end{split}
\end{equation}
The redundancy $R$ always has a positive value for a biased probability distribution, and we can say that learning is the detection of redundancy.

\subsection{Detection of redundancy}
\label{sec: Detection of the redundancy}
The next question is how our brain detects such redundancy. Let us consider the coded information which has redundancy. For example, let us consider the following binary code:
\begin{equation}
    C = 111000111111000111000111000000 \, .
\end{equation}
Then we can compress this information with following replacement: 
\begin{equation}
\begin{split}
   000 &\to 0 \\
   111 &\to 1 \\
   C &\to C^\prime = 1011010100\, .
\end{split}
\end{equation}
Even for general cases, using some methods, we can compress the information. Then, where does the redundancy go? The answer is that redundancy is embedded in how we compress the information. If there is redundancy in the information, we can find some way to compress it, and the redundancy corresponds to the rule of how we compress it. Summarizing the statement of Sec.\,\ref{sec: Consideration about human learning} up to now, the learning is the discovery of the correlation, the correlation makes the probability bias, the probability bias makes the redundancy, and the redundancy can be compressed by the proper rule for the compression. We can conclude that learning is the discovery of the rule for the compression of information.

\subsection{Compression of information}
\label{sec: The compression of information}
The next question is how our brain compresses the information. This is nothing but the theme of this paper, and our answer is explained as follows. The ultimate motivation of the human brain is to survive, and for this purpose, the human brain tries to record incoming information as much as possible. However, the amount of incoming information is very huge, and the human brain is limited, so it can record only compressed information. For example, when we see the image of an apple, millions of our optic neurons carry the information to our brain, but we only remember very abstract information such as ``the image was an apple.", and we forget the details of the image. This is lossy compression, and there is the loss of information by forgetting the detail of the information. In this process, the human brain tries to reduce the amount of information to remember and also tries to reduce the loss of information, as a survival strategy. We develop the quantitative evaluation of these amounts of information and find that the ideal data compression can be achieved by minimizing these values. In the following paper, we discuss the evaluation of these values and how the compression is achieved by considering the NN model as a toy model of our brain.

%%%%%%%%%%%%%%%%%%%%%%%%%%%%%%%%%%%%
%%%%%%%%%%%  model       %%%%%%%%%%%
%%%%%%%%%%%%%%%%%%%%%%%%%%%%%%%%%%%%
\section{The model of neural network}
\label{sec: The model of neural network}
In this section, we define a new NN model as a toy model of the human brain to evaluate the amount of information to remember and the loss of the information. Our goal is to derive these quantities, and we do not care about the actual implementation of the model. 

\begin{figure}[t]
    \centering
    \includegraphics[width=120mm]{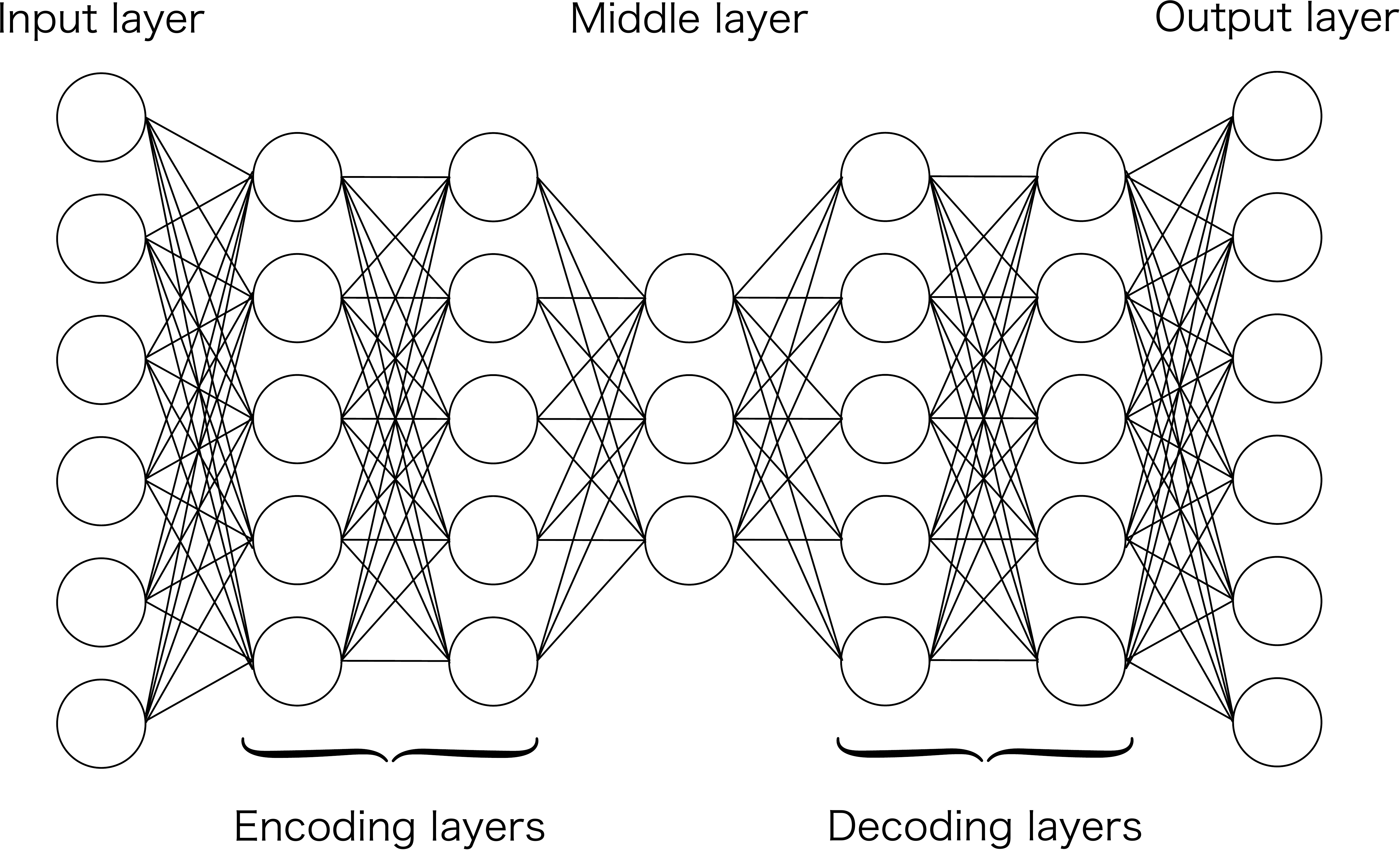}
    \caption{\small \sl The conceptual diagram of autoencoder NN model. The autoencoder has input layer, encoding layer, middle layer, decoding layers and output layer. The dimension of the middle layer is less than that of the input layer. The output of the NN tries to reproduce the input layer from the middle layer with less dimension.}
    \label{fig: autoencoder}
\end{figure}

First, as an example of the NN model which compresses the input data without ad hoc treatment, let us explain the overview of autoencoder NN\,\cite{doi:10.1126/science.1127647}. The autoencoder has the input layer, encoding layers, middle layer, decoding layers, and output layer as shown in Fig.\,\ref{fig: autoencoder}. Here, the dimension of the middle layer is less than that of the input layer, and the dimension of the output layer is the same as that of the input layer. The loss function is defined as the difference between the input and output, so the NN tries to restore the input data from the reduced information in the middle layer. We can consider this system as the compression and decompression of the input data, and the middle layer can extract the features of the input data.

\begin{figure}[t]
    \centering
    \includegraphics[width=130mm]{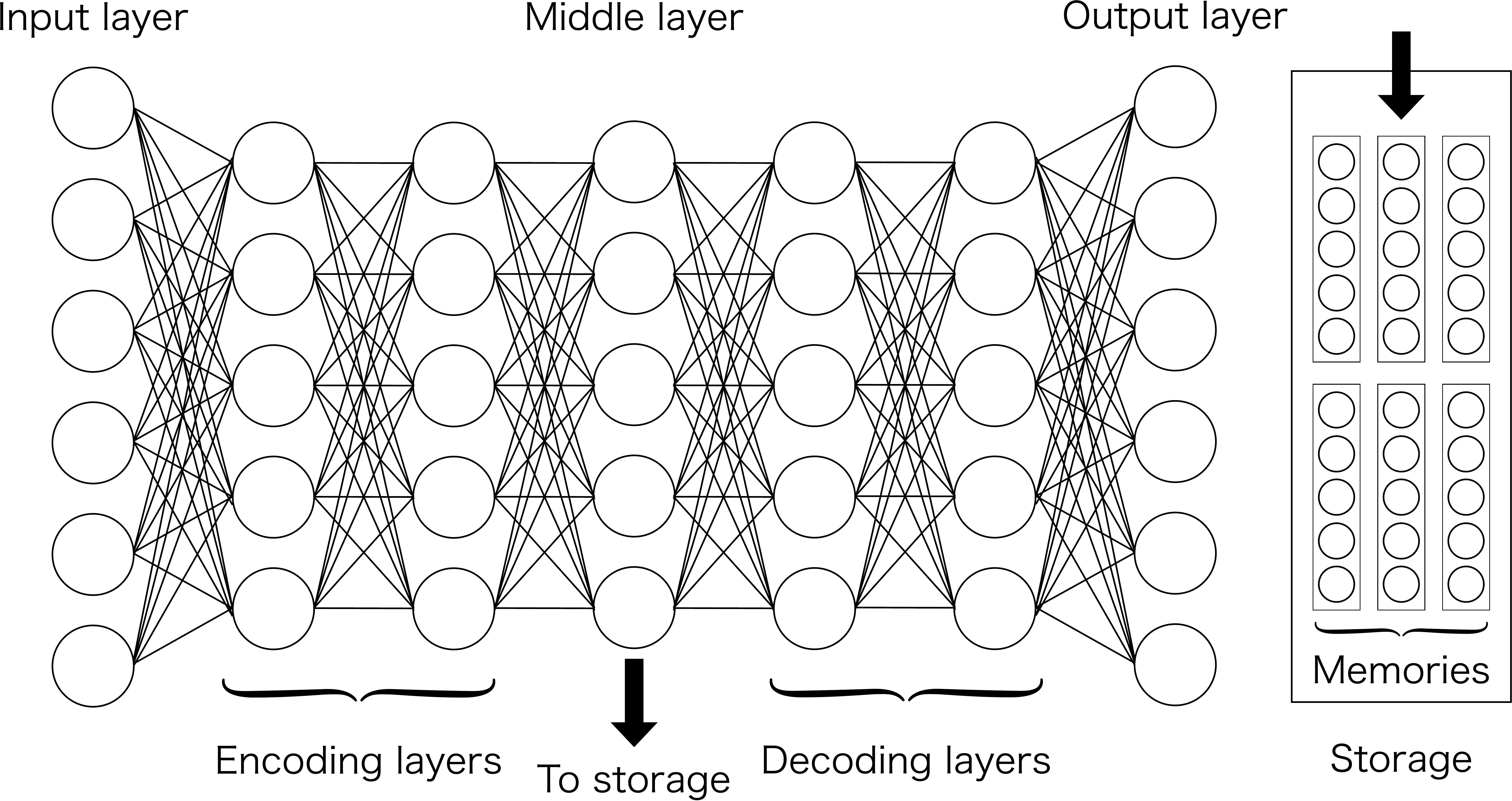}
    \caption{\small \sl The conceptual diagram of our NN model. Different from the usual autoencoder model, the dimension of the middle layer does not need to be less than that of the input layer. Instead, we record the value of the middle layer to the storage every time when new input enters and the calculation is done. We call the output from the middle layer the memory.}
    \label{fig: newNN}
\end{figure}

However, this model has two problems to achieve our goal. First, there is no guideline to choose the dimension of the middle layer, so we choose it looking at the result. We have to step out of such ad hoc treatment. The second problem is there is no evaluation of the amount of information to remember, and the NN only tries to reduce the loss of the information. We need a new system to treat the amount of information to remember and the loss of the information on the same ground.   

We suggest a new NN model to solve these problems. For the first problem, we do not constrain the dimension of the middle layer if it is large enough to contain the features of the input. For the second problem, we record the value of variables in the middle layer to the storage every time a new input enters, and use them to evaluate the amount of information to remember. We call the value of the middle layer the memory in the following paper, and we need to calculate the self-information of the memory. The conceptual diagram for this model is shown in Fig.\,\ref{fig: newNN}. We will discuss how the self-information of the memory is evaluated in Sec.\,\ref{sec: The self-information and the loss function}. 

To clarify the situation, we explain it more concretely. Suppose the dimension of the input layer is 4, the dimension of the middle layer is 3, and the dimension of the input layer is 4. The input $E$ has the form of a vector with dimension $4$, and it is written as  
\begin{equation}
    E =(e_1,e_2,e_3,e_4) \, .
\end{equation}
The output from the middle layer, or the memory $M$, has the form of a vector with dimension $3$, and it is written as 
\begin{equation}
    M =(m_1,m_2,m_3) \, ,
\end{equation}
where $M$ is calculated from the input $E$ and we record $M$ to the storage. The output $O$ has the form of a vector with dimension $4$, and it is written as
\begin{equation}
    O =(o_1,o_2,o_3,o_4) \, ,
\end{equation}
where $O$ is calculated from the memory $M$. If a new input $E^\prime$ enters, the NN calculate a new memory $M^\prime$ and a new output $O^\prime$, and we record $M^\prime$ to the storage. Repeating this process, we have a lot of memories in the storage. Here, we consider that $e_i$ and $m_i$ take discrete values, namely, $E$ and $M$ are discrete vectors. This is a very important assumption, and we will mention it in Sec.\,\ref{sec: The self-information and the loss function}.

\section{The self-information and the loss function}
\label{sec: The self-information and the loss function}
Our goal is to formularize the amount of information to remember and the loss of information. The amount of information is usually discussed in the framework of self-information. In this framework, we mention how to define the amount of information of the memory and the loss of information. Finally, we derive the loss function which can be used generally.

\subsection{The self-information}
\label{sec: The information entropy}
Suppose that $E$ is the event that occurs with the probability $P(E)$. We can consider that the rare event has large information, and we can formularize the self-information $I(E)$ as 
\begin{equation}
\label{eq: ie}
    I(E) = - \log{P(E)} \, .
\end{equation}
Let us consider $E$ as the input of the NN defined in Sec.\,\ref{sec: The model of neural network}. Here, we do not restrict the type of $E$. We consider that $E$ corresponds to the raw data of our optic nerve or auditory nerve when we see or hear something. For example, if $E$ is the event of seeing a cat, we enter $E$ as a form of pixel image data of a cat. Strictly speaking, our nerves carry the chemical substance, and the number of them must be discrete not continuous. Hence, we consider $E$ as a discrete vector. For the usual event, we cannot define $P(E)$ because there are so many types of inputs and the same input does not occur twice. However, if we could prepare the infinite number of input data sets, the same events occur many times for the discrete vector, and we can define $P(E)$ well. In this case, $P(E)$ takes a very small but non-zero value, and in the following, we take $P(E)$ as this meaning.  

Next, we will derive the amount of information to remember. We defined the memory as the value of variables in the middle layer in Sec.\,\ref{sec: The model of neural network}, and let us consider the the amount of information of the memory $M$. We also take $M$ as a discrete vector. By the definition of the self-information, we can write the self-information of $M$ as 
\begin{equation}
    I(M) = - \log{P(M)} \, .
\end{equation}
Here, the question is what $P(M)$ is. This is the probability of getting the memory $M$, and we can define $P(M)$ by using the recorded memories as 
\begin{equation}
\label{eq: memory probability}
    P(M) = \frac{n(M)}{N} \, ,
\end{equation}
where $n(M)$ is the number of the memory $M$ recorded up to now and $N$ is the total number of memories in the storage. We emphasize that the NN may record the same memory from different inputs because the memory is a discrete vector, and we count how many times the same memory $M$ was recorded, which corresponds to $n(M)$. However, if the parameter space of the memory is much larger than $N$, $n(M)$ equals 0 usually, and it becomes difficult to define $P(M)$ well. The prescription for this problem will be mentioned in Sec.\,\ref{sec: The loss function}.   

Next, let us consider the loss of information along with the compression. When $E$ enters as the input, we have all the information of $E$, but after recording $M$, we only have the information that $M$ is recorded. In this situation, there is a possibility that $M$ is recorded from the different input $E^\prime$ and we cannot tell which was the actual input only from the memory $M$. For example, suppose drawing one card from playing cards, such as king of hearts. Then, suppose that we only remember that the suit of the card was the heart and forget the number. For this case, we lose the information of number and this loss of information $L$ can be evaluated as
\begin{equation}
\label{eq: cards}
\begin{split}
    L &= -\log{P(\mathrm{King\; of\; hearts}|\mathrm{Hearts})}\, \\
      & = -\log{(1/13)} \, ,
\end{split}    
\end{equation}
where $P(A|B)$ denote the conditional probability where $A$ happens under the condition of $B$. Generalizing this statement to our NN model, the loss of information $L$ when $E$ enters under the condition that $M$ is recorded is written as
\begin{equation}
    L = -\log{P(E|M)}.
\end{equation}
Here, we can observe interesting equations from these quantities. From the Bayes' theorem, we can write the probability of input $E$ as
\begin{equation}
    P(E) = P(M)P(E|M) \, .
\end{equation}
Then taking the log of both sides and converting the sign, we get 
\begin{equation}
\label{eq: equation}
\begin{split}
    I(E) & =  \big(- \log{P(M)} - \log{P(E|M)}\big) \\
    &= I(M) + L \, .
\end{split}
\end{equation}
This corresponds to the conservation of self-information, which means that some portion of the input information will be embedded in the memory and the other portion of the input information will be lost.

\subsection{The loss function}
\label{sec: The loss function}
Next, we explain how to estimate $I(M)$ and $L$ using the NN defined in Sec.\,\ref{sec: The model of neural network}. Then, we will see this estimation is nothing but the loss function that we want to derive.

\begin{figure}[t]
    \centering
    \includegraphics[width=110mm]{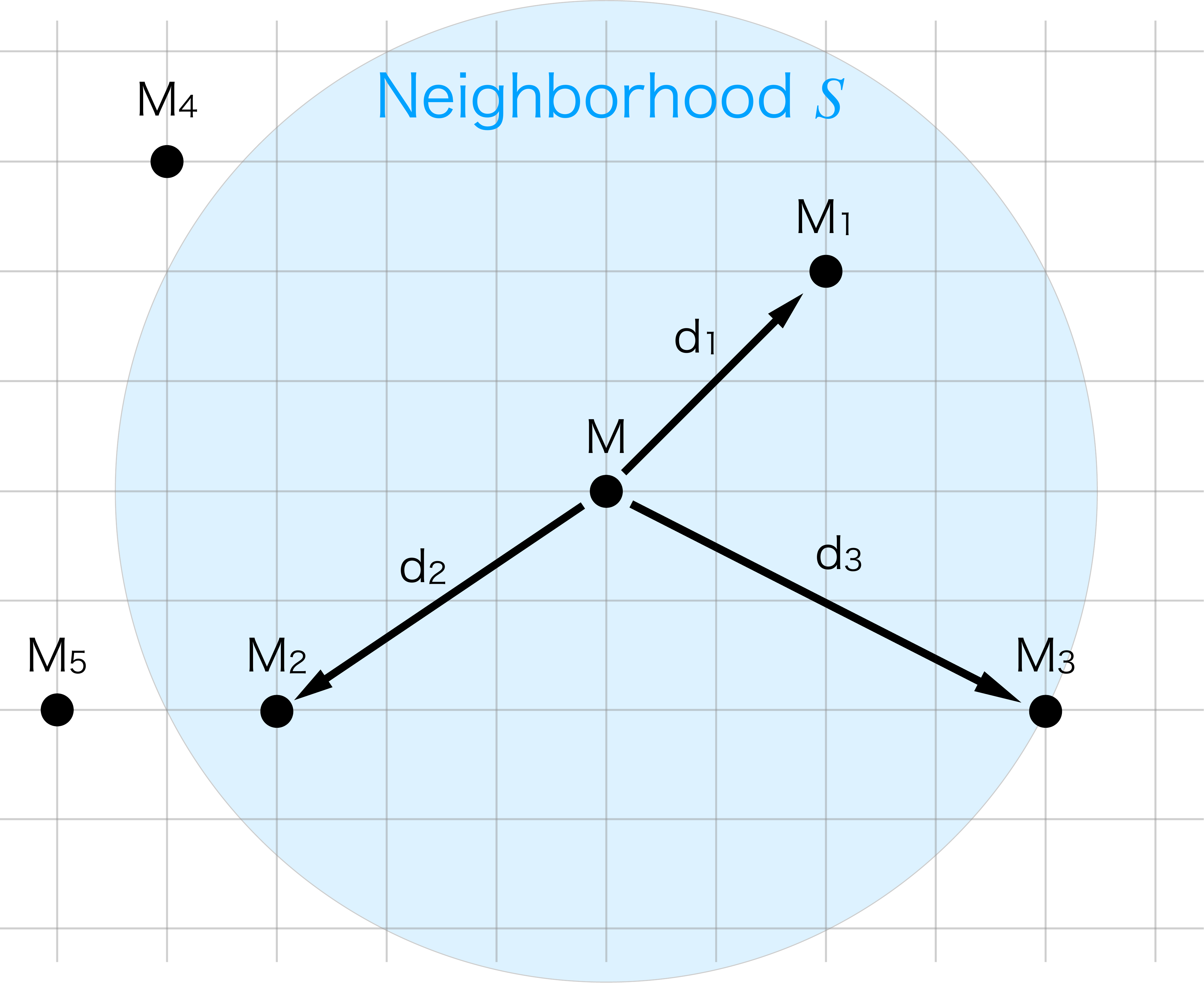}
    \caption{\small \sl The conceptual diagram of neighborhood $S$ from the memory $M$. The memories can be recorded as the lattice points, and black dots denote the memories in the storage. This diagram corresponds to the case of taking $n=3$ in Eq.\,(\ref{eq: dn}), and we take the average among the lattice points in the blue circle to calculate  $\mathcal{P}(M)$.}
    \label{fig: memory}
\end{figure}

Let us start from the estimation of $I(M)$. For this purpose, we need to estimate $n(M)$. However, we have the problem that $n(M)$ tends to be 0 when the parameter space of the memory is large. For example, when the middle layer has 100 dimensions and each node takes a binary value, the number of cases is $2^{100}$. This value must be larger than $N$, and exactly the same memory will not appear. To solve this problem, we add the condition that a similar memory should express similar data, and take the average among the near region. From this condition, we can consider $P(M)\simeq P(M^\prime)$ if $M\simeq M^\prime$. Then we define the estimation of $P(M)$ as 

\begin{equation}
\label{eq: memory}
\begin{split}
    \mathcal{P}(M) &\equiv  \mathrm{avg}\big(P(X)|_{X\in S}\big) \\
    & = \frac{\mathrm{avg}\big(n(X)|_{X\in S}\big)}{N} \, ,
\end{split}
\end{equation}
where $\mathcal{P}(M)$ denotes the estimation of $P(M)$, $\mathrm{avg}(\cdot)$ is the function to take the average of the argument, and $S$ is the neighborhood of the memory $M$. By taking the average among the neighborhood $S$, we can get a non-zero value for $\mathcal{P}(M)$ if we set $S$ as large enough to contain a non-zero number of memories. Here, the problem is how to define $S$. One suggestion is as follows. Let us name the memories in the storage as $M_1, M_2, \dots, M_N$, then we calculate the distance between a new memory $M$ and a recorded memory $M_i$ as
\begin{equation}
    d_i = \sqrt{(M-M_i)^2} \, .
\end{equation}
Next, we rename $d_i$, arranging in ascending order as $d_1\leq d_2\leq \dots  \leq d_N$ is satisfied. Then we define $S$ using the distance between memories as 
\begin{equation}
\label{eq: dn}
    S = \left\{X\big{|}(X-M)^2 \leq (d_{n})^2 \right\} \, ,
\end{equation}
where $n$ is the number of the memories that are contained in the neighborhood $S$, and we can choose an arbitrary number for $n$. By doing this, $\mathcal{P}(M)$ takes the average in the neighborhood within the radius $d_n$. We show the conceptual image of the neighborhood $S$ in Fig.\,\ref{fig: memory}. This suggestion for the definition of $S$ is just an example, and it will be improved in future work.  

Next, we mention the estimation $L$ using this NN. First, let us discuss the case of Eq.\,(\ref{eq: cards}) more deeply. The conditional probabilities under the condition of getting hearts are written as 
\begin{equation}
\label{eq: conditional}
\begin{split}
    P(\mathrm{Ace\; of\; spades}&|\mathrm{Hearts}) = 0 \, ,\\
    &\vdots    \\ 
    P(\mathrm{King\; of\; spades}&|\mathrm{Hearts}) = 0 \, ,\\
    P(\mathrm{Ace\; of\; hearts}&|\mathrm{Hearts}) = 1/13 \, ,\\
    & \vdots    \\ 
    P(\mathrm{King\; of\; hearts}&|\mathrm{Hearts}) = 1/13 \, ,\\    P(\mathrm{Ace\; of\; diamonds}&|\mathrm{Hearts}) = 0 \, ,\\
    & \vdots    \\ 
    P(\mathrm{King\; of\; diamonds}&|\mathrm{Hearts}) = 0 \, ,\\
    P(\mathrm{Ace\; of\; clubs}&|\mathrm{Hearts}) = 0 \, ,\\
    & \vdots \\
    P(\mathrm{King\; of\; clubs}&|\mathrm{Hearts}) = 0 \, .
\end{split}
\end{equation}
This is the list of $P(X|\mathrm{Hearts})$ where $X$ can take any kind of playing cards, and we can think that $P(\mathrm{King\; of\; hearts}|\mathrm{Hearts})$ corresponds to the case with $X = \mathrm{King\; of\; hearts}$. By the analogy of this case, we first consider $P(X|M)$ to get $P(E|M)$ where $X$ can be any type of input data. Hence, the output of the NN tries to estimate the probability distribution $P(X|M)$, and let us call this output $\mathcal{P}(X|M)$. We give an example of such an output. Let us take the input $E$ as binary data set with dimension $3$, such as $(0,1,0)$. Through the NN, this input is compressed into the memory $M$ and the output predicts the probability of the input from the compressed information $M$. The output has $3$ nodes and each node tries to predict the probability of the corresponding node of input data to have the value 1. For example, if the output is $(0.1,0.9,0.2)$, this output predicts that the first node of the input takes 1 with 10\,\%, the second node of the input takes 1 with 90\,\%, and the third node of the input takes 1 with 20\,\%. In this case, the $\mathcal{P}(E|M)$ is calculated as 
\begin{equation}
\begin{split}
 \mathcal{P}(E|M) = (1-0.1)\times0.9\times(1-0.2) \, .
\end{split}
\end{equation}
Though we just explained the case with dimension $3$, we can generalize this for any number of input dimensions. Moreover, any type of input data can be recast into binary data, and it means that we can always use this method in principle.

Finally, we set the loss function $\mathcal{L}(E)$ to be the sum of the estimation of $I(M)$ and $L$ as 
\begin{equation}
\label{eq: loss}
\mathcal{L}(E) \equiv - \log{\mathcal{P}(M)}  - \log{\mathcal{P}(E|M)}\, .
\end{equation}

\subsection{The meaning of the loss function}
To discuss the meaning of the loss function, we need to mention the important property of the information entropy. Suppose $P(X)$ and $Q(X)$ is the probability distribution about $X$ and
\begin{equation}
    \sum_X P(X) = \sum_X Q(X) = 1
\end{equation}
is satisfied. In this case, we can write the following inequality as 
\begin{equation}
    -\sum_X P(X) \log{P(X)} \leq -\sum_X P(X) \log{Q(X)} \, ,
\end{equation}
where left-hand side corresponds to the information entropy, or the expectation value of self-information for $P(X)$. This inequality always holds for any probability distribution $P(X)$ and $Q(X)$, and the equal sign holds only when $P(X) = Q(X)$ is satisfied. 

Using this inequality for the true probability distribution of input $P(E) = P(M)P(E|M)$ and the prediction of the probability distribution of input $\mathcal{P}(E) = \mathcal{P}(M)\mathcal{P}(E|M)$, we get  
\begin{equation}
        \sum_E P(E) (-\log{P(M)} -\log{P(E|M)}) \leq  \sum_E P(E) (-\log{\mathcal{P}(M)} -\log{\mathcal{P}(E|M)}) \, .
\end{equation}
Here, the left-hand side corresponds to the expectation value of $I(E)$, and the right-hand side corresponds to the expectation value of $\mathcal{L}(E)$. Then we can write the following inequality as 
\begin{equation}
\label{eq: IL}
       I(E)|_\mathrm{exp} \leq \mathcal{L}(E)|_\mathrm{exp} \, ,
\end{equation}
where $\cdot|_\mathrm{exp}$ means the expectation value of the function. This inequality has a very important meaning. As the learning proceeds, the loss function $\mathcal{L}(E)$ gets closer to the true self-information of the input $I(E)$, and the minimum point corresponds to $I(E)$. Using this loss function, we can estimate the actual self-information $I(E)$ without knowing the true probability distribution $P(E)$.

\section{The role of memory and the hidden variables}
\label{sec: The role of the memory and its probability distribution}
In Sec.\,\ref{sec: Consideration about human learning}, we mentioned that learning is the detection of redundancy. In this section, we discuss how the redundancy is compressed, and we also consider the role of the memory.

\subsection{The redundancy}
Although we briefly explained about the redundancy in Sec.\,\ref{sec: Probability bias and redundancy}, we will explain the more general and complicated cases here. Suppose that we take the animal images as the input of the NN, and each image is expressed as the pixel binary data. Let us discuss what is redundancy in this case. Taking the number of the pixels of the input data as $l$, the number of the cases of the input data is $2^l$. If there is no probability bias, each data will appear completely randomly, and we can calculate the maximum of the information entropy $H_\mathrm{max}$ as  
\begin{equation}
\begin{split}
        H_\mathrm{max} &= -\log{2^{-l}} \, ,
\end{split}
\end{equation}
Among the random inputs, only a very tiny parameter region can be considered to be an animal image. Hence, the input of animal image have very peaky probability bias on the parameter space of the input, and the expectation value of self-information is written as
\begin{equation}
\begin{split}
        I(E)|_\mathrm{exp} &= -\sum_{E} P(E)\log{P(E)} \\ 
        &= -\log{2^{-l}} - R\, ,
\end{split}
\end{equation}
where $E$ denotes the input of animal image, and $R$ is the redundancy defined in Eq.\,(\ref{eq: redundancy}). From the eq.\,(\ref{eq: IL}), the NN compresses $R$ as the learning proceed. 

\subsection{The hidden variables}
\label{sec: The hidden variable}
The input of the animal image has $l$ variables, and these variables are strongly correlated and form probability bias. We can consider this situation as follows. There are hidden variables such as the kind of the animal, the size of the animal, and the posture of the animal, and the input data is decided by these hidden variables. Here, the dimension of the hidden variables is less than $l$. Considering the meaning of the redundancy $R$ from this perspective, we find there are two types of redundancy. The first is the redundancy from the mapping of the hidden variables to input data. For example, if the hidden variables specify the kind of animal as a cat, there must be mapping to input data that looks like a cat, and it makes the correlation between the input variables. The second is the redundancy from the prior probability distribution of hidden variables. For example, if the probability of the hidden variables which denote a cat is much higher than that of a kangaroo, it also can be a source of redundancy.   

\subsection{The role of memory}
\label{sec: The role of the memory}
\begin{figure}[t]
    \centering
    \includegraphics[width=170mm]{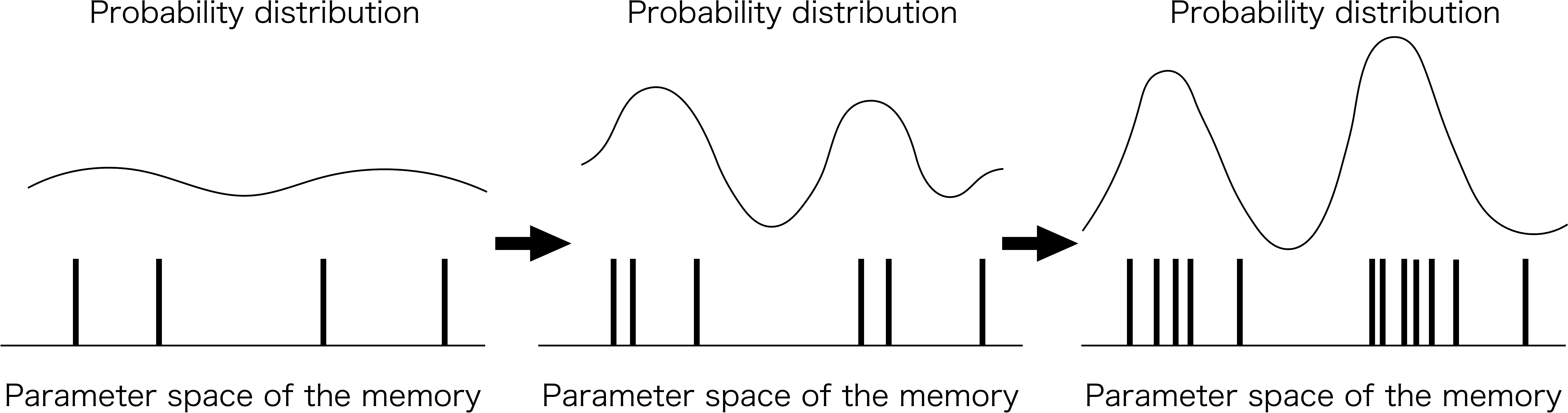}
    \caption{\small \sl The conceptual diagram which shows how the memories are recorded and how the probability distribution of the memory is updated. Each vertical black line denotes each memory.}
    \label{fig: memory distribution}
\end{figure}

The NN tries to reduce $-\log\mathcal{P}(M)$, which means the NN tries to increase $\mathcal{P}(M)$, and from Eq.\,(\ref{eq: memory}), the NN tries to record the memories as near as possible. We show the conceptual diagram in Fig.\,\ref{fig: memory distribution}. Namely, the memory tries to record the information with fewer variables, and this is nothing but the hidden variables. From this consideration, we guess that the redundancy from mapping of the hidden variables to input data is embedded in the encoding layer, and the redundancy from the prior probability distribution of the hidden variables is embedded in the probability distribution of the memory. The hidden variables are considered to have abstract information, such as ``this is a cat" or ``this is sitting", and recording the memory corresponds to recording such abstract information. This is very similar to the human brain which only remembers abstract information and forgets trivial information. 

\begin{figure}[t]
    \centering
    \includegraphics[width=90mm]{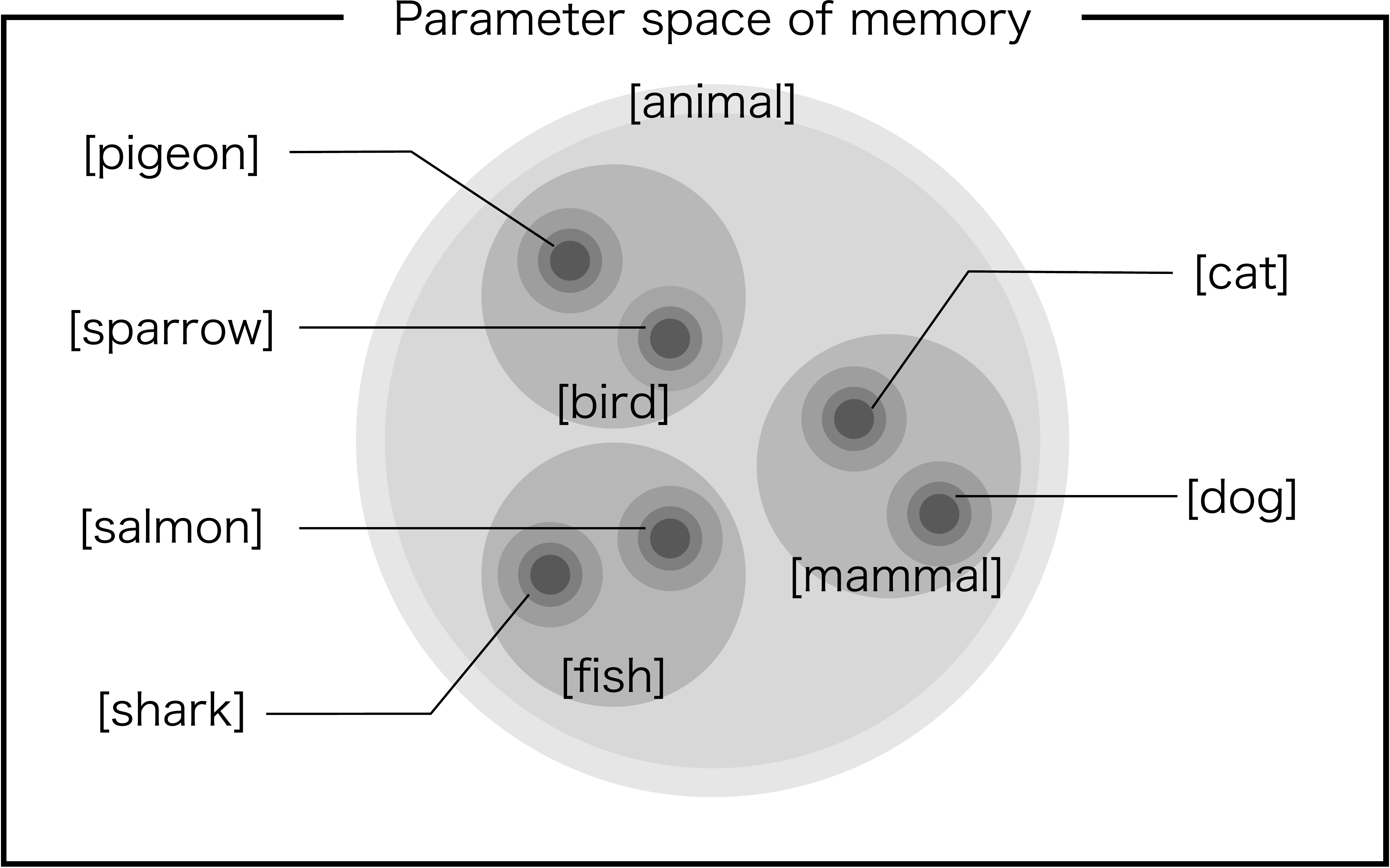}
    \caption{\small \sl The conceptual diagram of the probability distribution in the parameter space of memory. We considered the case of learning the image of animals, and a darker region denote higher probability, in other words, memories are distributed with a higher density in a darker region. The peaks of the probability distribution correspond to the species of animals.}
    \label{fig: animal}
\end{figure}

Next, let us discuss the probability distribution of the memory. Suppose the learning of the animal image, there considered to be many hidden variables, such as ``having hair", ``having a tail", or ``having slippery skin". Among the combination of these hidden variables, some combinations will appear frequently. For example, the combination of ``having slippery skin" and ``having a streamlined shape" means the characteristics of fish, and other animals also have such a combination of the hidden variables. These combinations have a higher probability than random combinations, and it means that there emerge some peaks of the probability distribution in the parameter space of memory. We can consider that each peak corresponds to the species of animals. If the learning proceed more, there emerge minor peaks which correspond to a detailed classification of the animals. We illustrate this situation in Fig.\,\ref{fig: animal}. In this figure, the darker region has a higher probability distribution, which means the density of the recorded memories is higher in such region. The important point is that such a structure of the probability distribution of memory automatically emerges without ad hoc treatment. We guess that this is exactly the way how the human brain remembers and learns things. 

\subsection{The memory as input}
\label{sec: The memory type object}
Next, let us consider what happens if we set the memories as input. In Sec.\,\ref{sec: The role of the memory}, we mentioned that memories also have their probability distribution. This means that memories are also controlled by other hidden variables, and we can compress the information again. Repeating this process many times, we can find deeper hidden variables which express more abstract information. We can also set different types of memories as input simultaneously, then the NN will learn the correlation between different types of information. For example, if we enter the memories from voice data of the name of animals and the memories from image data of animals simultaneously, the NN will learn the name of animals. The essence of learning is finding the hidden variables by compressing the input data, and the NN will learn more deeply by repeating this process.

\section{The uncertainty of the solution}
\label{sec: The uncertainty of the solution}
Here, we mention the uncertainty of the solution which minimizes this loss function. The self-information of the memory and the loss of information are inextricably linked, and there are several types of realization to obtain the minimum of the loss function whether the NN tends to record information as much as possible or abandon information as much as possible. To control this uncertainty, we suggest a new loss function as 
\begin{equation}
\begin{split}
    \mathcal{L}(E;\alpha,\beta) &=  - (1+\alpha)\log{\mathcal{P}(M)} - (1+\beta)\log{\mathcal{P}(E|M)} \, 
\end{split}
\end{equation}
where $\alpha$ and $\beta$ is a small non-negative constant. In the following, we demonstrate how this loss function works by changing the value of $\alpha$ and $\beta$, using a very simple example. To help understand how this loss function works, we consider two-dimensional binary input data and name them as 
\begin{gather}
    E_1 \equiv (0,0),\; E_2 \equiv (0,1),\; E_3 \equiv (1,0),\; E_4 \equiv (1,1). 
\end{gather}
We also set the memory to have the shape of two-dimensional binary data and name them as 
\begin{gather}
    M_1 \equiv (0,0),\; M_2 \equiv (0,1),\; M_3 \equiv (1,0),\; M_4 \equiv (1,1).
\end{gather}
Then, for instance, let us consider the case mentioned in Sec.\,\ref{sec: Correlation and probability bias} and set the probability as same as Eq.(\ref{eq: case 2})
\begin{gather}
    P(E_1) =0.6 ,\;P(E_2) =0.1,\;P(E_3) =0.1,\; P(E_4) =0.2 \, .
\end{gather}
The expectation value of the self-information of input data is calculated as
\begin{equation}
    \begin{split}
    I(E)|_\mathrm{exp} &= -\sum_{i=1}^4 P(E_i)\log{P(E_i)} \\ 
    &= 1.088\cdots \, .
\end{split}
\end{equation}
Next, we consider when the loss function takes its minimum, changing the value of $\alpha$ and $\beta$. Here, there are only four types of memories, and we do not need to average over the neighborhood to calculate $\mathcal{P}(M)$.

\subsection{The case with $\alpha$=0, $\beta$ =0.01 }
Let us take $\mathcal{L}(E;0,0.01)$ as the loss function. The minimum of the loss function is achieved when the encoding layers work as 
\begin{equation}
\begin{split}
    E_1 \to M_1, \\
    E_2 \to M_2, \\
    E_3 \to M_3, \\
    E_4 \to M_4,     
\end{split}
\end{equation}
and the decoding layers work as
\begin{equation}
\begin{split}
    M_1 \to (p_1, p_2) = (0,0), \\
    M_2 \to (p_1, p_2) = (0,1), \\
    M_3 \to (p_1, p_2) = (1,0), \\
    M_4 \to (p_1, p_2) = (1,1).     
\end{split}
\end{equation}
Then the expectation value of the loss function is
\begin{equation}
\begin{split}
    \mathcal{L}(E,0,0.01)|_\mathrm{exp} =& -0.6\times\left[(1+0)\log{0.6}+(1+0.01)\log{1}\right] \\
    & -0.1\times\left[(1+0)\log{0.1}+(1+0.01)\log{1}\right] \\
    & -0.1\times\left[(1+0)\log{0.1}+(1+0.01)\log{1}\right] \\
    & -0.2\times\left[(1+0)\log{0.2}+(1+0.01)\log{1}\right] \\
    =& 1.088 \cdots \, .
\end{split}
\end{equation}

\subsection{The case with $\alpha$=0.01, $\beta$ =0 }
Next, let us take $\mathcal{L}(E;0.01,0)$ as the loss function. The minimum of the loss function is achieved when the encoding layers work as 
\begin{equation}
\begin{split}
    E_1 \to M_1, \\
    E_2 \to M_1, \\
    E_3 \to M_2, \\
    E_4 \to M_2,     
\end{split}
\end{equation}
and the decoding layers work as
\begin{equation}
\begin{split}
    M_1 \to (p_1, p_2) = (0,1/7), \\
    M_2 \to (p_1, p_2) = (1,2/3), 
\end{split}
\end{equation}
Then the expectation value of the loss function is
\begin{equation}
\begin{split}
    \mathcal{L}(E,0.01,0)|_\mathrm{exp} =& -0.6\times\left[(1+0.01)\log{0.7}+(1+0)\log{6/7}\right] \\
    & -0.1\times\left[(1+0.01)\log{0.7}+(1+0)\log{1/7}\right] \\
    & -0.1\times\left[(1+0.01)\log{0.3}+(1+0)\log{1/3}\right] \\
    & -0.2\times\left[(1+0.01)\log{0.3}+(1+0)\log{2/3}\right] \\
    =& 1.095 \cdots \, .
\end{split}
\end{equation}

\subsection{The case with $\alpha$=0.2, $\beta$ =0 }
Next, let us take $\mathcal{L}(E;0.2,0)$ as the loss function. The minimum of the loss function is achieved when the encoding layers work as 
\begin{equation}
\begin{split}
    E_1 \to M_1, \\
    E_2 \to M_1, \\
    E_3 \to M_1, \\
    E_4 \to M_2,     
\end{split}
\end{equation}
and the decoding layers work as
\begin{equation}
\begin{split}
    M_1 \to (p_1, p_2) &= (1/8,1/8), \\
    M_2 \to (p_1, p_2) &= (1,1), 
\end{split}
\end{equation}
Then the expectation value of the loss function is
\begin{equation}
\begin{split}
    \mathcal{L}(E,0.2,0)|_\mathrm{exp} =& -0.6\times\left[(1+0.2)\log{0.8}+(1+0)\log{49/64}\right] \\
    & -0.1\times\left[(1+0.2)\log{0.8}+(1+0)\log{7/64}\right] \\
    & -0.1\times\left[(1+0.2)\log{0.8}+(1+0)\log{7/64}\right] \\
    & -0.2\times\left[(1+0.2)\log{0.2}+(1+0)\log{1}\right] \\
    =& 1.203 \cdots \, .
\end{split}
\end{equation}

\subsection{The case with $\alpha$=0.5, $\beta$ =0 }
Next, let us take $\mathcal{L}(E;0.5,0)$ as the loss function. The minimum of the loss function is achieved when the encoding layers work as 
\begin{equation}
\begin{split}
    E_1 \to M_1, \\
    E_2 \to M_1, \\
    E_3 \to M_1, \\
    E_4 \to M_1,     
\end{split}
\end{equation}
and the decoding layers work as
\begin{equation}
\begin{split}
    M_1 \to (p_1, p_2) &= (0.3,0.3), \\
\end{split}
\end{equation}
In this case, there are only four type of memories, and we do not need to average over the neighborhood to calculate $\mathcal{P}(M)$. Then the expectation value of the loss function is
\begin{equation}
\begin{split}
    \mathcal{L}(E,0.5,0)|_\mathrm{exp} =& -0.6\times\left[(1+0.5)\log{1}+(1+0)\log{0.49}\right] \\
    & -0.1\times\left[(1+0.5)\log{1}+(1+0)\log{0.21}\right] \\
    & -0.1\times\left[(1+0.5)\log{1}+(1+0)\log{0.21}\right] \\
    & -0.2\times\left[(1+0.5)\log{1}+(1+0)\log{0.09}\right] \\
    =& 1.221 \cdots \, .
\end{split}
\end{equation}

\subsection{Properties of the loss functions}
From these observations, we can state as follows. If $\alpha$ is zero, the NN tries to record as much information as possible, and if $\beta$ is zero, the NN tries to abandon as much information as possible. As $\alpha$ becomes bigger, the NN starts to reduce the information to remember, and abandons the information of the correlation between two nodes. For the case with $\alpha = 0.01, \beta = 0$ and $\alpha = 0.2, \beta = 0$, the NN uses two memories $M_1$ and $M_2$. Considering the case mentioned in Sec.\,\ref{sec: Correlation and probability bias}, the memory $M_2$ corresponds to the concept which connects the word cat and the visual image of a cat.

\section{Discussion}
\label{sec: Discussion}
First, we discuss how we solved the problem mentioned in Sec.\,\ref{sec: Introduction}. We wondered about the trend of the current study of the NN, in which people focus on how to get good results by ad hoc treatment. People are trying to improve the NN model, the loss function, or the style of input data. However, considering the human brain, it does not need such treatments and can learn everything automatically in daily life. There must be a unified theory that controls the human brain, and we do not need ad hoc treatment under such a theory. This theory is nothing but the loss function we defined in Eq.(\ref{eq: loss}). Let us mention that our theory includes previous ad hoc treatments. For example, convolutional NN is doing coarse-graining, and this can be considered as some kind of compression of the input data. We know the fact coarse-graining does not lose so much information, and this fact is implemented in the NN ad hoc. The situation for other good-looking models are more or less same. People consider the property of the input data and try to use this property and extract the features of input data by improving the NN models. This corresponds to teaching the NN how to compress the input data ad hoc. Our theory offers a unified treatment for any kind of input data.

Next, we discuss the memory. We mentioned that the memory expresses the hidden variable in Sec.\,\ref{sec: The hidden variable}, but the realization is not unique. The probability distribution of the memory would be different if we clean up all the memories and restart the learning. Moreover, if we change the definition of the neighborhood $S$ considered in Sec.\,\ref{sec: The information entropy} or change the value of $\alpha$ and $\beta$ considered in Sec.\,\ref{sec: The uncertainty of the solution}, the probability distribution of the memory will change more drastically. There is a lot of uncertainty, and we need to improve the theory about the hidden variables. Let us also mention about the treatment of old memories. As the learning proceeds, many memories are recorded in the storage, but the memories recorded in the early stage of learning do not follow the desired probability distribution. To improve the convergence of the calculation, we should remove such memories from the storage. This also remains as future work.

Finally, we discuss the similarity between this NN and the human brain. This NN is suggested as a toy model of the human brain, and we do not need ad hoc treatment for the different inputs. We can use the same type of NN for a different types of input data, and this is very similar to our brain which can adapt to any kind of problem. This is achieved only by minimizing the loss function, which is defined as the sum of the self-information to remember and the loss of information. This situation amazingly agrees with the free-energy principle\,\cite{Friston10}. The free-energy principle is the theory in cognitive science proposed by Karl Friston, in which the cognition, learning, and action of the organism are determined by minimizing the function called free-energy. Our loss function corresponds to this free-energy, and it is formularized in a calculable way. To calculate the self-information to remember, we introduced the concept of memory. This is the core of our theory, and it has many similarities with our actual memory. The memory denotes compressed information, and it can express abstract information. The human brain also remembers only compressed and abstract information. In the paragraph above, we discussed the benefit to forget old memories, and this situation is also similar to our brain, as the human brain tends to forget old memories which are not often referred to.

\section{Conclusion}
\label{sec: Conclusion}
In this paper, we proposed a unified theory of learning. Learning is the discovery of the correlation between the information, and this can be achieved through the compression of the information. To explain how this compression is carried out, we treat a NN model as a toy model of the human brain and derived a new loss function. In this process, we proposed the concept of memory, and this is the core of our theory. The loss function is expressed as the sum of the self-information of the memory and the loss of information. This reflects the self-information of the input data, and we can estimate it without knowing the actual probability distribution of the input data only by calculating the minimum of the loss function. To estimate the self-information of the memory, the NN refers to the recorded memories. Usually, the density of the memories is very low, and we need to refer to similar memories. 

We can think that the correlation of the information is caused by the hidden variables, and the memory expresses these hidden variables. The hidden variables also have their probability distribution, and it emerges as the density distribution of the memories. The peaks of the density distribution correspond to the features of the input data. The memories have their probability distribution, and it means we can find deeper hidden variables by setting the memories as input. We can also set the different tyeps of memories as input simultaneously, then the NN will find the correlation between different data. This means the NN can learn the abstract concept, such as the meaning of the words, by setting the memories of audio data and image data as input. Deeper hidden variables express more abstract information, and the reputation of this process is nothing but how the human brain learns.

We also found many common features in this NN model and the human brain. We do not need ad hoc treatment for the different types of inputs, and this is the same as the human brain which learns by itself. This is achieved only by minimizing the loss function which denotes the sum of the information to remember and the loss of the information, and this corresponds to the free-energy theorem which is considered to be the unified theory of the human brain. The memory of this NN works very similarly to our actual memory, and it can express very abstract information. The similar information is recorded as a similar memory. We also mentioned the benefit to forget old memories.    

This work has a lot of implications for a very wide range of fields, and it can be applied to any kind of data analysis. However, there will be many difficulties in the implementation of this work in the actual NN system. We only offered the definition of the loss function, and we have not checked the NN numerically. To improve the convergence of the calculation, we will need a lot of study. We also offered the concepts of memory and hidden variables, but these concepts have so many theoretical uncertainties. They are totally new concepts and we do not know the properties of them at all. They will be understood through the calculation of concrete examples. We need not only experimental approach but also theoretical approach to the NN. 

%%%%%%%%%%%%%%%%%%%%%%%%%%%%%%%%%%%%%%%
%%%%%%%%%%% Acknowledgments %%%%%%%%%%%
%%%%%%%%%%%%%%%%%%%%%%%%%%%%%%%%%%%%%%%
\section*{Acknowledgments}
The author would like to thank Prof. Haruhiro Katayose for the useful discussion and suggestions. This work is supported in part by the JSPS KAKENHI Grant No.\,20H00160.

%%%%%%%%%%%%%%%%%%%%%%%%%%%%%%%%%%
%%%%%%%%%%% References %%%%%%%%%%%
%%%%%%%%%%%%%%%%%%%%%%%%%%%%%%%%%%
\bibliographystyle{apsrev4-1}
\bibliography{refs}

\end{document}